\documentclass[11pt]{article}

\usepackage[preprint]{acl}

\usepackage{times}
\usepackage{latexsym}

\usepackage[T1]{fontenc}
\usepackage[utf8]{inputenc}

\usepackage{microtype}

\usepackage{inconsolata}

\usepackage{graphicx}

\usepackage{tcolorbox}
\tcbuselibrary{breakable}
\usepackage{fontawesome5} % For beautiful icons
\usepackage{xcolor} % For color definitions
\tcbuselibrary{skins}
\tcbuselibrary{listingsutf8}
\usepackage[edges]{forest}

\usepackage{tabularx, booktabs, xcolor, colortbl}
\usepackage{pifont}   %
\usepackage{xstring}  %

\usepackage{multirow}
\usepackage{hyperref}
\usepackage{enumitem}

\usepackage[tikz]{bclogo}
\usepackage[framemethod=tikz]{mdframed}

\usepackage[export]{adjustbox}

\newcommand{\tstyle}[1]{\underline{\textit{#1}}}

\definecolor{rqblue}{HTML}{E8F1FA}
\definecolor{findinggreen}{HTML}{E8F9E8}
\definecolor{findingbrown}{HTML}{D4A37F}

\newtcolorbox{findingbox}[1][]{
    breakable,
    enhanced,
    sharp corners,
    boxrule=0pt,
    colback=findingbrown!20,
    colframe=findingbrown!20,
    frame hidden,
    borderline west={2pt}{0pt}{findingbrown},
    left=6pt,
    right=6pt,
    top=4pt,
    bottom=4pt,
    before skip=10pt,
    after skip=10pt,
    fontupper=\linespread{1.0}\selectfont,
    #1
}

\newcommand{\researchfinding}[1]{%
  \begin{findingbox}
    \textcolor{findingbrown}{\faLightbulb[regular]}~#1
  \end{findingbox}
}

\definecolor{findingbrown}{HTML}{D4A37F}
\definecolor{training_process}{HTML}{1A75D5}
\definecolor{model}{HTML}{D3450A}
\definecolor{special_behavior}{HTML}{29A198}

\tikzset{%
    every node/.style={font=\tiny},
    parent/.style =          {align=center,text width=1.5cm,rounded corners=3pt, line width=0.3mm, fill=gray!15,draw=gray!80},
    child/.style =           {align=center,text width=2.0cm,rounded corners=3pt, fill=blue!10,draw=blue!80,line width=0.3mm},
    grandchild/.style =      {align=center,text width=2cm,rounded corners=3pt},
    greatgrandchild/.style = {align=center,text width=1.5cm,rounded corners=3pt},
    greatgrandchild2/.style = {align=center,text width=1.5cm,rounded corners=3pt},    
    referenceblock/.style =  {align=center,text width=1.5cm,rounded corners=2pt},
    pretrain/.style =           {align=center,text width=2.0cm,rounded corners=3pt, fill=training_process!15,draw=training_process!80,line width=0.3mm},   
    pretrain_work/.style =           {align=center, text width=9.5cm,rounded corners=3pt, fill=training_process!15,draw=training_process!0,line width=0.3mm},  
    template/.style =           {align=center,text width=2.0cm,rounded corners=3pt, fill=model!15,draw=model!80,line width=0.3mm},   
    template_work/.style =           {align=center,text width=9.5cm,rounded corners=3pt, fill=model!15,draw=model!0,line width=0.3mm},    
    answer/.style =           {align=center,text width=2.0cm,rounded corners=3pt, fill= special_behavior!15,draw= special_behavior!80,line width=0.3mm},   
    answer_work/.style =           {align=center,text width=9.5cm,rounded corners=3pt, fill= special_behavior!15,draw= special_behavior!0,line width=0.3mm},
}

\hypersetup{
    linkcolor=red,
linktocpage}

\title{Towards a Mechanistic Understanding of Large Reasoning Models: \\A Survey of Training, Inference, and Failures}

\author{Yi Hu\textsuperscript{1} 
\quad Jiaqi Gu\textsuperscript{1} 
\quad Ruxin Wang\textsuperscript{1} 
\quad Zijun Yao\textsuperscript{2}
\quad Hao Peng\textsuperscript{2}\\
\quad {\bf Xiaobao Wu\textsuperscript{3}}
\quad {\bf Jianhui Chen\textsuperscript{2}} 
\quad {\bf Muhan Zhang\textsuperscript{1, \dag}} 
\quad {\bf Liangming Pan\textsuperscript{1,\dag}}\\
  \textsuperscript{1}Peking University\quad
  \textsuperscript{2}Tsinghua University\quad\\
  \textsuperscript{3}Nanyang Technological University
}

\begin{document}
\maketitle

\renewcommand{\thefootnote}{\fnsymbol{footnote}} % 用符号编号脚注
% \footnotetext[1]{Equal contribution.}
\footnotetext[2]{Corresponding author. \\Correspondence: \texttt{huyi2002@stu.pku.edu.cn, \{muhan, liangmingpan\}@pku.edu.cn}}

\begin{abstract}
Reinforcement learning (RL) has catalyzed the emergence of Large Reasoning Models (LRMs) that have pushed reasoning capabilities to new heights. 
% capable of complex, multi-step reasoning. 
% Beyond the excitement surrounding their performance, 
While their performance has garnered significant excitement, exploring the internal mechanisms driving these behaviors has become an equally critical research frontier. This paper provides a comprehensive survey of the mechanistic understanding of LRMs, organizing recent findings into three core dimensions: 1) training dynamics, 2) reasoning mechanisms, and 3) unintended behaviors. By synthesizing these insights, we aim to bridge the gap between black-box performance and mechanistic transparency. Finally, we discuss under-explored challenges to outline a roadmap for future mechanistic studies, including the need for applied interpretability, improved methodologies, and a unified theoretical framework. 

\textbf{Our Project: }\href{https://github.com/AheadOFpotato/Awesome-LRM-Mechanisms}{\faGithub \ Awesome-LRM-Mechanisms}
% By synthesizing these insights, we aim to bridge the gap between black-box performance and mechanistic transparency in the next generation of reasoning models. Finally, we highlight XXX under-explored open research questions to outline a roadmap for future mechanistic studies. 

% . First, we dissect training dynamics, examining the distinct roles of Supervised Fine-Tuning (SFT) in expanding capability boundaries versus RL in compressing reasoning paths, alongside the emergence of "aha moments." Second, we analyze reasoning mechanisms, linking behavioral patterns like backtracking and self-verification to internal representations and attention circuit dynamics. Finally, we investigate unintended behaviors, uncovering the mechanistic roots of hallucinations, unfaithful chains of thought, and overthinking. By synthesizing these insights, we aim to bridge the gap between black-box performance and mechanistic transparency in the next generation of reasoning models.
\end{abstract}

\begin{figure*}[th]
    \centering
    \includegraphics[width=\linewidth]{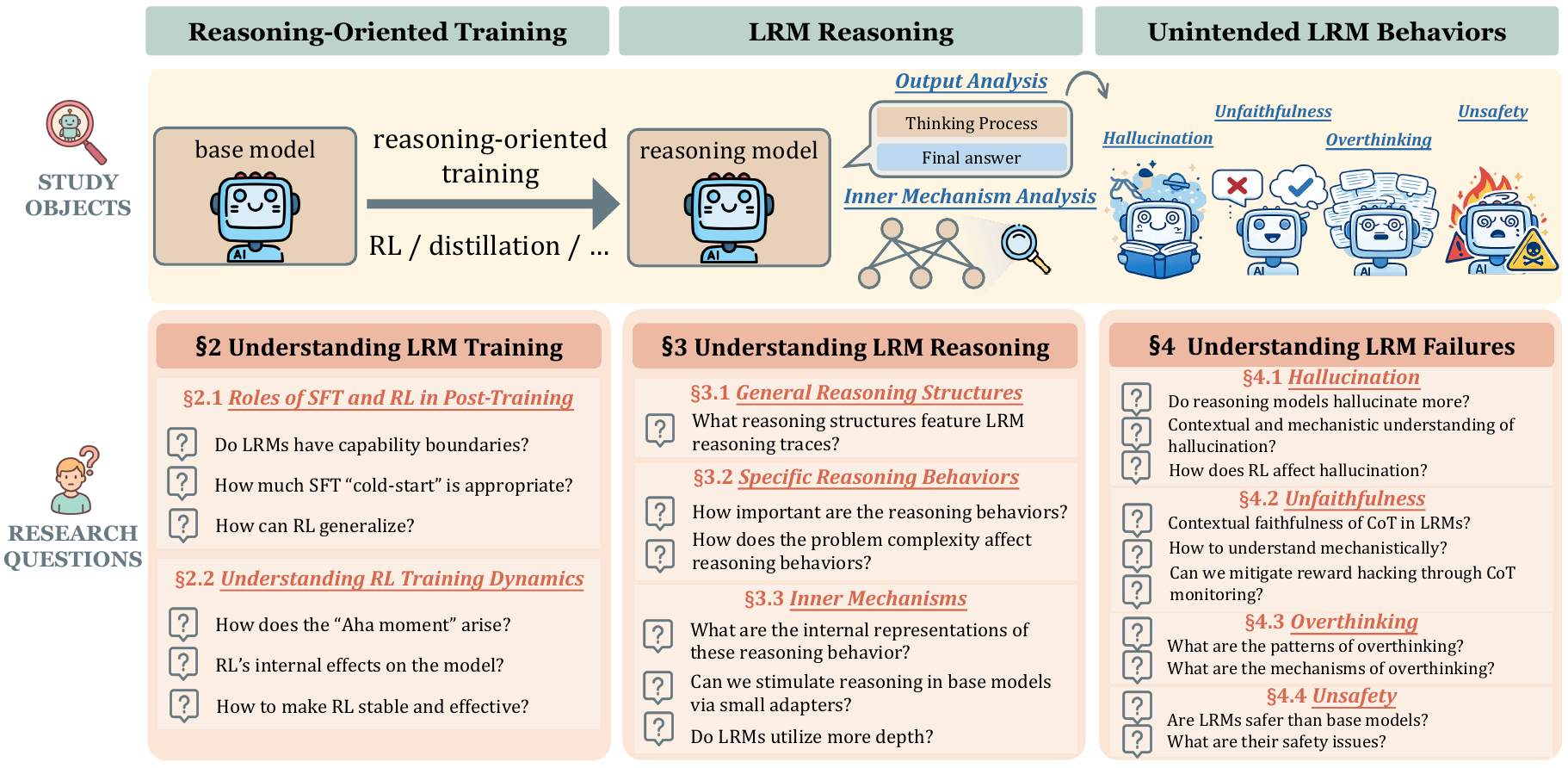}
    % \caption{The taxonomy of our survey. According to research objects, we study the reasoning-oriented training process in Sec~\ref{sec:training}, LRM reasoning behaviors in Sec~\ref{sec:model} and LRM unintended behaviors in Sec~\ref{sec:special_behavior}.}
    \caption{Taxonomy of mechanistic research on LRMs. We organize existing studies into three core dimensions: reasoning-oriented training (Sec~\ref{sec:training}), reasoning mechanisms (Sec~\ref{sec:model}), and unintended behaviors (Sec~\ref{sec:special_behavior}). Within each dimension, we synthesize recent findings based on the key research questions being investigated in the literature.}
    \vspace{-0.4cm}
    \label{fig:taxonomy}
\end{figure*}

\section{Introduction}
The past few years have witnessed remarkable progress in the reasoning capabilities of large language models (LLMs). Recently, reinforcement learning (RL) has emerged as a transformative paradigm for incentivizing complex reasoning, giving rise to advanced large reasoning models (LRMs)~\citep{deepSeek_R1,o1}. These models demonstrate exceptional performance across a wide range of domains, including mathematics, coding, and logic. Notable research~\citep{deepSeek_R1} has shown that RL from verifiable rewards (RLVR)~\citep{deepSeek_R1,Tulu32024} training can elicit intriguing emergent reasoning behaviors, such as extended reasoning chains and self-reflection.

Despite these impressive advances, LRMs largely remain ``black boxes''. Many fundamental questions remain unanswered, including: How does the role of RL differ from that of supervised fine-tuning (SFT)? What structural properties define LRM reasoning, and what are the internal mechanisms that drive their unique behaviors? Moreover, what are the root causes of unintended behaviors, such as hallucinations, unfaithfulness, and overthinking? This lack of transparency has spurred a growing interest in mechanistic research, aimed at uncovering the underlying processes that enable these models to perform complex reasoning.

We provide a comprehensive survey of the burgeoning field of mechanistic research on LRMs. From the perspective of the research object, as shown in Figure~\ref{fig:taxonomy}, we organize work studying the reasoning-oriented training process, LRM reasoning behaviors and LRM unintended behaviors:

\begin{enumerate}[
  topsep=0pt,      % 列表顶部与上文间距
  partopsep=0pt,   % 当列表开始新段落时的额外间距
  itemsep=0pt,     % 列表项之间的间距
  parsep=0pt,      % 段落内的间距（与itemsep类似）
  leftmargin=*,    % 左边界对齐
]
\item \textbf{Reasoning-Oriented Training Process} (\S\ref{sec:training}): This section examines the mechanisms behind the training processes that specifically target reasoning capabilities. We begin by dissecting the complementary roles of SFT and RL (\S\ref{ssec:training/post_training}), and examine key training dynamics in RL, such as how ``aha moments'' emerge and how internal representations evolve during training (\S\ref{ssec:training/how_rl_shape}).

\item \textbf{LRM Reasoning} (\S\ref{sec:model}): We delve into the mechanisms underlying LRM reasoning, analyzing both their outputs and internal representations. This section explores the general structural features of LRM reasoning traces (\S\ref{ssec:model/General Reasoning Structures}), key behaviors like self-reflection (\S\ref{ssec:model/Specific Reasoning Behaviors}), and the inner mechanisms underlying these behaviors (\S\ref{ssec:model/Internal Mechanisms}).

\item \textbf{Unintended LRM Behaviors} (\S\ref{sec:special_behavior}): We further examine the side effects of LRMs, exploring behavioral patterns and internal mechanisms associated with typical unintended behaviors, such as hallucinations (\S\ref{ssec:hallucination}), unfaithful chains of thought (CoT) (\S\ref{ssec:faithfulness}), overthinking (\S\ref{ssec:overthink}), and unsafety (\S\ref{ssec:unsafety}).
\end{enumerate}

\paragraph{Contribution and Uniqueness.} Our survey distinguishes itself by focusing specifically on the \textit{mechanistic understanding} of LRMs, a topic that has received limited attention in existing literature. While several surveys provide general overviews of large reasoning models and RL techniques~\citep{zhang2025100,li2025system,zhang2025survey,Survey_Reasoning_Model}, these surveys do not delve deeply into the underlying mechanisms driving LRM reasoning. In particular, \citet{long_cot_survey} explores long CoT reasoning but primarily focuses on the behavioral characteristics of CoT outputs, with little attention to the inner mechanisms. Furthermore, there are surveys investigating methods to mitigate overthinking~\citep{feng2025efficient,sui2025stop}, their focus is on efficient reasoning techniques rather than the mechanisms behind overthinking. To the best of our knowledge, our work is the first to comprehensively survey the mechanisms of LRMs, offering a more detailed and in-depth analysis of the training processes, reasoning behaviors, and unintended outcomes.

\section{Understanding LRM Training}
\label{sec:training}

Investigating how a reasoning model is trained into existence is our first question. To understand LRM training, we will first dissect the distinct roles of two post-training methods, \textit{Supervised Fine-tuning} (SFT) and \textit{Reinforcement Learning} (RL) (\S\ref{ssec:training/post_training}), then dive into the training dynamics of RL (\S\ref{ssec:training/how_rl_shape}).

\subsection{Roles of SFT and RL in Post-Training}
\label{ssec:training/post_training}

DeepSeek-R1 \cite{deepseekai2025deepseekr1incentivizingreasoningcapability} as a key pioneer of the reasoning model, demonstrates RL’s vital role in training reasoning models. However, they also find that RL alone suffers from problems including slow training, unstructured formats, and language mixing. They show that a cold-start with SFT fixes these problems and improves performance, establishing SFT+RL as the dominant post-training recipe for today’s LRMs. Although the SFT+RL paradigm is now widely used, the respective roles of SFT and RL in post-training remain to be explored, and answering this question would unravel many mysteries about LRM training.

% Although RL plays a crucial role in the emergence of reasoning ability, the present post-training paradigm remains an SFT + RL pipeline. It is essential to understand and compare the roles played by the different algorithms throughout the whole process and to analyze how they jointly shape the reasoning capability of the model. 

\paragraph{SFT explores, RL compresses: do reasoning models have capability boundaries?}

% To supply usable data and rewards for large-scale RL training, the idea of Reinforcement Learning with Verifiable Rewards (RLVR, \citet{Tülu 3: Pushing Frontiers in Open Language Model Post-Training}) was proposed and adopted in the training of Reasoning models such as Deepseek-R1.

Despite the huge success of LRMs, \citet{yue2025does} points out that RLVR~\citep{Tulu32024} do not truly enhance reasoning performance beyond base models, instead, they compress the model’s output space, boosting pass@1. The base model’s pass rate catches up in pass@\textit{k} tests at large \textit{k} values, implying that the reasoning model merely uncovers latent abilities already present in the base model.
Based on this finding, follow-up work probe deeper and reveal that SFT is what truly expands the model’s exploratory paths, whereas RL training compresses them, cutting the variety of possible answers and lowering output entropy \cite{matsutani2025rlsqueezessftexpands, wu2025invisibleleash, li2025tracing}. Studies further conclude that the performance gains from RLVR training come from this entropy drop, imposing an entropy-linked ceiling on attainable capability \cite{cui2025entropymechanismreinforcementlearning}. 
From a mechanistic perspective, \citet{park2025thinking} perform circuit analyses to explain this phenomenon: during SFT and distillation, the model sprouts a cohort of new attention heads that inject reasoning capabilities, whereas GRPO activates far fewer heads and follows an iterative activate-evaluate-adjust dynamic.

\paragraph{SFT learns, RL repairs: how much SFT cold-start is appropriate?}

Under the above findings, SFT appears to be the core contributor. However, research shows that while SFT can learn and extend reasoning patterns, unlike RL, it also brings out-of-distribution (OOD) performance drops \cite{chu&zhai2025sftmemorizes}. Studies find that RL after SFT can partly repair these side-effects and offer mechanistic explanations. % for how SFT causes and RL repairs these problems.
\citet{hu&cai2025howllmlearn} argues, through external analysis, that SFT partially breaks up the sparse reasoning concept network and thereby induces forgetting, yet the network structure remains largely intact; RL can then restore a well-connected concept network after SFT. 
\citet{jin&luan2025rlfinetuning,jin2025rlisneither} find that RL can, yet only partially, recover the OOD drop caused by SFT. Besides, OOD performance is tightly linked to the orientation of the dominant singular vector, and RL repairs the orientation shift introduced by SFT, thereby restoring OOD accuracy. However, once SFT collapses into overfitting, RL can no longer restore OOD ability completely.

Building on these findings, how to schedule SFT and RL, whether to interleave them, and whether a unified framework can be designed that fuses them have become open research questions; we summarize SFT-RL integration work in Appendix~\ref{ssec:appendix1}. %  move 3.2 to appendix, linked by this paragraph

\paragraph{SFT memories, RL organizes: how can we make RL exhibit generalization?}

Although numerous studies have concluded that RL does not truly enhance model capability, \citet{liu2025prorl} shows that post-RL models can produce new solutions absent from the base model. How can these two contradictory experimental conclusions be reconciled? Recent work suggests that RL can push the model’s capability frontier outward only when SFT has provided basic skills. In controllable synthetic reasoning tasks, RL conducted after SFT with atomic-skill data exhibit OOD generalization , while RL conducted after SFT on entire reasoning traces exhibits the same poor generalization results seen in prior research.\cite{yuan&chen2025fromf(x), cheng2025fromatomic}. Furthermore, these studies indicate that RL shifts the model’s error patterns toward atomic-task errors, implying that RL can indeed help organize the model’s reasoning process.

% By crafting fully controllable synthetic reasoning tasks, they find that if the model is first given SFT on atomic reasoning skills and then receives RL on composite reasoning data, its final performance on OOD composite tasks rises significantly. In contrast, training with arbitrary data via SFT, or doing SFT directly on composite-reasoning tasks and then appending RL, both fail to yield OOD generalization. \cite{yuan&chen2025fromf(x), cheng2025fromatomic} They further find that RL shifts the model’s error patterns toward atomic-task errors, indicating that RL can indeed teach the model to organize its reasoning process. % which means that the prevailing view that RL cannot extend a model’s reasoning ability does not stem from an inherent limitation of RL itself, but from the improper training data.

\emph{Mid-training} proposed by \citet{wang2025octothinker} , which finds that appropriate training before RL can improve RL effectiveness, aligns with the above conclusions. \citet{zhang2025ontheinterplay} notes that the core task of SFT is to prepare the model for RL by providing foundational atomic skills, while post-RL training refines the model's performance within the capability frontier established by SFT.

%%%%% Optional
% \researchfinding{SFT provides the foundational capabilities, while RL builds on these atomic skills to enhance reasoning and generalization.}

%On the Interplay of Pre-Training, Mid-Training, and RL on Reasoning Language Models
%OctoThinker: Mid-training Incentivizes Reinforcement Learning Scaling

\subsection{Understanding RL Training Dynamics}
\label{ssec:training/how_rl_shape}

Studies above treat the RL-training stage as an undifferentiated whole and explore its effects. The finer-grained training dynamics within this stage remain largely unexplored.

\paragraph{Understanding two-stage training process: how does the Aha moment arise?}

After tracking changes of RL training metrics, studies split the training process into two stages~\citep{wang2025emergent, hu&cai2025howllmlearn}. Model outputs first shrink in stage one then lengthen in stage two, alongside atomic-skill fragments rapidly acquired in stage one while the global planning links slowly built in stage two. The aha moment emerges as the model masters the use of planning tokens during link construction manifesting as the sudden acquisition of reasoning and reflection capabilities needed to solve the corresponding problems. %  Regarding the speed difference between the two stages, the latter study proposed that the reasoning graph formed by LRM outputs is subject to strict degree constraints which cause the sluggish progress in second stage.
Furthermore, \citet{yao2025thedebate} offers a theoretical analysis of this two-stage dynamics. During stage one, RL overwhelmingly samples already-explored tokens rather than optimal ones. High-reward tokens' probabilities will quickly rise while the optimal one's remain flat. In stage two, with high-reward tokens already saturated, the low-probability optimal ones are finally sampled after prolonged exploration and eventually receive high probabilities.

\paragraph{What internal effects does RL have on the model?}
A line of research focuses on how RL training affects the model internally.
Regarding internal activations, research shows that online RL can alter activation magnitudes in the residual stream, increasing information flow flexibility and improving generalization beyond SFT~\cite{zhang2025reinforcement}. Regarding model weights, building on the previously identified effect that RL training mainly manifests as directional rotation of the singular-value vectors \cite{jin2025rlisneither}, \citet{he2025understandingpost-training} reveals through SVD methods that a near-uniform geometric scaling of singular values across layers and a highly consistent orthogonal transformations are applied to the left and right singular vectors of each matrix. More fine-grained studies of parameter dynamics during training have found that, the top singular subspace of the parameter-update matrix almost singly accounts for the gains in reasoning capability, and that this dominant subspace evolves linearly \cite{cai2025onpredictability}.  

% Building on this, their \emph{AlphaRL} exploits the rank-1 dynamics observed in the early phase to predict the final update matrix without running the full training, cutting RL training time dramatically while preserving most of the performance.

% \researchfinding{Online RL can alter activation magnitudes in residual stream, and the RL training process manifests primarily as rotational transformations of singular-value vectors, and its impact is highly consistent and rank-1 dominated.}

\researchfinding{Externally, RL training shows a two-stage pattern: basic capabilities are accumulated first, then reasoning ability emerges. Internally, RL modifies activation magnitudes and applies a rank-1, layer-consistent, linear rotational transformation to the dominant eigenvectors.}

\paragraph{Exploitation v.s. exploration: how to make RL stable and effective?}

During RL training, a core issue is maintaining the exploration-exploitation balance. Studies find that basic RL algorithms can easily lead to \textbf{policy entropy collapse}, and the performance gains in fact come solely from the entropy drop \cite{cui2025entropymechanismreinforcementlearning}. More critically, \citet{nguyen2025reasoning} shows that the reasoning path compression caused by entropy collapse simultaneously degrades LRM performance on questions outside the training distribution. %, resulting in a winner-take-all scenario.
To address entropy collapse and stabilize RL, numerous refinements have been proposed. Since they are loosely related to understanding RL training mechanisms, we provide a concise summary in  Appendix~\ref{ssec:appendix2}.
Notably, \citet{huang2025beyond} argues that conventional RLVR views improving LLM performance through an exploration–exploitation trade-off, rests on token-level entropy and thus misaligns with how LLMs actually operate. They propose measuring exploration and exploitation via \textit{hidden states}, uncovering a decoupling of the two processes and opening fresh avenues for refining RL algorithms.

% \citet{nguyen2025reasoning} studied the learning dynamics of RLVR and revealed negative interference in LLM reasoning. They found that learning to solve a subset of training problems can \textit{negatively affect others}, resulting in a winner-take-all scenario. In this situation, the model disproportionately improves on a limited subset of problems that the base model can already solve with high likelihood, while neglecting or even regressing on others. 

\researchfinding{At token level, the entropy collapse can make RL training ineffective or even counter-productive. Shifting to the hidden states perspective, we may instead jointly promote exploration and exploitation.}

\section{Understanding LRM Reasoning}
\label{sec:model}
Having explored how LRMs are trained, we shift our focus to the models themselves, involving systematically analyzing both the \textbf{general structures} (\S\ref{ssec:model/General Reasoning Structures}) and \textbf{specific behaviors} (\S\ref{ssec:model/Specific Reasoning Behaviors}) within reasoning traces, as well as uncovering the \textbf{internal mechanisms} underlying these patterns (\S\ref{ssec:model/Internal Mechanisms}).

\subsection{General Reasoning Structures}
\label{ssec:model/General Reasoning Structures}

Distinct from base models, LRMs generate reasoning chains with identifiable structural features. Recent research deconstruct these traces, from macro-level lifecycle descriptions to granular sentence-level analyses. 
At the macro level, \citet{marjanovic2025deepseek} identifies a cyclical process: starting with problem definition, models enter a blooming cycle of problem decomposition, followed by iterative reconstruction cycles for self-correction before reaching a final decision.
\citet{wang2025accuracydissectingmathematicalreasoning} partitions the reasoning process into functional blocks of plan execution, knowledge integration, and subproblem chains. These macro-phases are further refined by sentence-level analyses: \citet{bogdan2025thoughtanchorsllmreasoning,li2025understandingthinkingprocessreasoning} identify operational units including plan generation, uncertainty management, and further identify the transition matrix between them.

\paragraph{Topological structures.} 
% \HY{confirm the results are collective conclusions of cited papers, can be more concise here} 
Another line of research employs formal topological representations. \citet{zeng2025rejumptreejumprepresentationanalyzing,jiang2025makesgoodreasoningchain} reconstuct reasoning chains as trees structures via LLM annotations, revealing that LRMs exhibit more exploration and validation than base models, achieving better performance primarily through diverse solution paths rather than per-step accuracy. \citet{minegishi2025topologyreasoningunderstandinglarge,xiong2025mappingmindsllmsgraphbased} build graphs through clustering over reasoning steps, further validating that LRMs possess distinct structural properties including more recurrent cycles, larger graph diameters, and pronounced small-world characteristics, which correlate with model size, task difficulty, and performance.

\researchfinding{
% LRMs' reasoning chains often exhibit stronger exploration, diverse path generation, and the ability to recover from erroneous paths; high accuracy stems from generating diverse solutions rather than optimizing a single path.
LRMs’ reasoning structures are distinct from base models, with analyses spanning \textit{macro-level lifecycles}, \textit{sentence-level operational units}, and \textit{topological properties of tree and graph representations}.}

\subsection{Specific Reasoning Behaviors}
\label{ssec:model/Specific Reasoning Behaviors}
After reviewing the overall reasoning structures, we further study the intriguing specific behaviors emerging in LRMs and whether they are causally related to reasoning performance.

 % \paragraph{The pivotal role of reasoning patterns.}
 % At the sentence level, \citet{bogdan2025thoughtanchorsllmreasoning} identifies eight distinct reasoning behaviors, discovering that "thought anchors"—such as plan generation, uncertainty management, and self-correction\HY{I remember only 2 behaviors are mentioned, confirm here}—exert a causal influence on subsequent steps via attention mechanisms, thereby significantly enhancing accuracy in complex tasks. From a structural perspective, \cite{jiang2025makesgoodreasoningchain} models CoT as a hierarchical tree representation, further emphasizing the importance of behaviors such as exploration and backtracking\HY{we do not write about structures any more in this subsection}. Distinct from these structural analyses, \cite{gandhi2025cognitivebehaviorsenableselfimproving} examines the efficacy of RL across different base models and finds that models naturally possessing initial cognitive behaviors like verification and backtracking exhibit significant performance gains during RL, whereas models lacking these traits encounter difficulties in improvement, \HY{add something like confirming the causal relationship between the cognitive behaviors with model performance}.
\paragraph{Critical behavioral primitives.} 
Studies identify certain behavioral patterns as the primary drivers of reasoning performance gain. \citet{bogdan2025thoughtanchorsllmreasoning} identifies ``thought anchors'', including \emph{plan generation} and \emph{uncertainty management}, as the sentences most influential on the final answer distribution. Complementing this, \citet{gandhi2025cognitivebehaviorsenableselfimproving} highlights \emph{verification}, \emph{backtracking}, \emph{sub-goal setting}, and \emph{backward chaining} as the ``four habits'' of effective reasoners. Crucially, these behaviors are causally linked to training success: base models that naturally exhibit these patterns can effectively leverage RL and test-time compute to improve performance, whereas models lacking these primitives struggle to benefit from identical training.

\paragraph{The role of self-reflection and backtracking.}
Research on reflective behaviors offers contrasting views. While some argue that reflection prevents reasoning collapse \cite{yang2025understandingahamomentsexternal}, others contend that it is often superficial and fail to improve outcomes \citep{liu2025oatzero}. Bridging these views, \citet{kang2025trymattersrevisitingrole} analyzes reflection from both inference and training perspectives, suggesting that while reflection during inference is largely confirmatory and rarely alters the final output, including reflective CoTs in training data increases the ``first-attempt accuracy'', boosting the overall performance. Moreover, \citet{cai2025backtrackingenoughexploringinterplay} shows that longer reasoning chains with frequent backtracking lead to more stable RL training, and harder problems with larger search space need the inclusion of data with more backtracks during SFT.

\researchfinding{LRMs' performance is driven by key behavioral primitives. While self-reflection mainly serves a confirmatory role during inference, its inclusion in training data is crucial for improving first-attempt accuracy and internalizing search strategies.}

\paragraph{How does the problem complexity affect reasoning behaviors?}
Recent studies have uncovered a tight coupling between model behavior and task complexity. \citet{yang2025understandingahamomentsexternal} observes that LRMs can distinguish problem complexity within their early layers and dynamically modulate the depth of their reflective behaviors accordingly. However, this calibration is often imperfect. \citet{shojaee2025illusionthinkingunderstandingstrengths} finds that while reasoning effort initially increases with complexity, it eventually declines even when a sufficient token budget is available, suggesting a limitation in the models' ability to apply consistent algorithmic reasoning across scales. Furthermore, \citet{palod2025performativethinkingbrittlecorrelation} identifies that the correlation is brittle, demonstrating that trace length often reflects a problem's distributional proximity to training data rather than its inherent computational complexity. We will further discuss the relationship between CoT length and task complexity, reasoning performance in Sec~\ref{ssec:overthink}.

\subsection{Internal Mechanisms}
\label{ssec:model/Internal Mechanisms}
After reviewing the general structures and specific behaviors, we will then dive deeper into the internal mechanisms driving these external patterns.

\paragraph{Internal representations of reasoning behaviors.}
% Some studies have conducted a holistic analysis of the reasoning patterns in models, identifying key reasoning patterns. For example, \citet{bogdan2025thoughtanchorsllmreasoning} identifies sentences related to Plan Generation and Uncertainty Management, which consistently receive the most attention through Receiver heads. Other studies focus on the origins and manipulation of model reasoning behaviors. \citet{venhoff2025understandingreasoningthinkinglanguage, venhoff2025basemodelsknowreason} identify steering vectors corresponding to different reasoning behaviors in the reasoning process, successfully controlling the model's reasoning behavior. Furthermore, in combination with Sparse Autoencoder (SAE) techniques,\citet{galichin2025icoveredbaseshere} have identified multiple interpretable features related to reasoning and used these features to manipulate the reasoning behaviors of LRMs.
Recent research utilizes sparse autoencoders (SAEs) and steering vectors to reveal that reasoning behaviors are encoded as interpretable and steerable directions in the model's activation space \citep{baek2025towards, galichin2025icoveredbaseshere, GoodfireBlog, venhoff2025understandingreasoningthinkinglanguage}. \citet{venhoff2025basemodelsknowreason} argues that base models already possess fundamental reasoning capabilities, while LRMs learn the structural strategy of \textit{when} to deploy them strategically. This deployment is managed by specific attention heads that prioritize key reasoning steps influencing the final answer ~\citep{bogdan2025thoughtanchorsllmreasoning,zhang2025reasoninganswerempiricalattentionbased}. LRMs also exhibit unique temporal and nonlinear dynamics: steering is most effective only after the initial problem formulation phase, and ``oversteering'' these features can paradoxically cause the model to revert to its original behavior \citep{GoodfireBlog}.

\paragraph{The mechanisms underlying reflection and backtracking.}
% Internal analyses suggest that self-reflection and backtracking are rooted in LRMs' latent representations.
Studies~\citep{venhoff2025basemodelsknowreason,yang2025understandingahamomentsexternal} reveal through linear probes that correctness information of model answers is encoded within specific layers, and is closely related to the model’s reflection behaviors. \citet{yan2025reflctrlcontrollingllmreflection,chang2025unveilinglatentdirectionsreflection} further extract steering vectors that control reflection. \citet{ward2025reasoningfinetuningrepurposeslatentrepresentations} suggests that latent directions for backtracking already exist in base models, implying that they inherently possess certain reasoning abilities. Post-training mainly reshapes and utilizes these existing representations rather than learning from scratch.

\paragraph{Can we stimulate reasoning behaviors in base models with small adapters?}
\citet{sinii2025steeringllmreasoningbiasonly,sinii2025smallvectorsbigeffects} have explored training hierarchical steering vectors to guide base models in reasoning, showing that the performance improvements induced by RL are distributed across the entire network instead of certain specific layers. The resulting steering vectors themselves exhibit strong interpretability. \citet{ward2025rank} trains a rank-1 adapter across all layers and identifies interpretable features in the adapter via SAEs, further demonstrating that a small number of parameters can effectively induce reasoning abilities.

\paragraph{Do LRMs utilize more depth?}
Research suggests that key layers for math reasoning are largely fixed after pre-training and remain invariant throughout post-training \citep{nepal2025layerimportancemathematicalreasoning}. Consequently, LRMs' effective depth closely matches that of their base models, indicating that improvements are driven by longer contexts rather than deeper per-token computation~\citep{hu2025affectseffectivedepthlarge}.

\researchfinding{LRM's reasoning behaviors are represented by interpretable and steerable directions in latent space; base models inherently possess these abilities, but RL-trained models learn when to activate them.}

\section{Understanding LRM Failures}
\label{sec:special_behavior}
RL enhances reasoning capabilities but also induces unintended effects, including \textbf{hallucination} (\S\ref{ssec:hallucination}), where models generate plausible yet incorrect content; \textbf{CoT unfaithfulness} (\S\ref{ssec:faithfulness}), where internal computations and CoT outputs diverge; \textbf{overthinking} (\S\ref{ssec:overthink}), where redundant reasoning chains degrade performance; and \textbf{unsafety} (\S\ref{ssec:unsafety}), where models show potentially harmful behaviors.

\subsection{Hallucination}
\label{ssec:hallucination}
Multi-step reasoning chains in LRMs introduces new vulnerabilities to hallucinations.

\paragraph{Do reasoning models hallucinate more?}
% Are Reasoning Models More Prone to Hallucination?
Recent evidence suggests that reasoning-oriented training pipelines can substantially affect hallucination behavior. \citet{are_reasoning_models-zijun} show that while complete post-training pipelines which combine SFT with RLVR can alleviate hallucination, incomplete pipelines, such as RL- or SFT-only approaches, tend to introduce more hallucinations.
However, \citet{lireasoning} indicates that RL often increases hallucinations, even with prior SFT. Furthermore, \citet{zhao2025test} shows that test-time scaling does not reliably improve factual accuracy.

\researchfinding{
  Growing evidence suggests reasoning models hallucinate more. However, there are debates whether models with complete post-training pipelnes hallucinate more.
}

\paragraph{Behavioral and mechanistic analysis of hallucination.}
% Research characterizes hallucinations through specific failure modes and internal mechanistic patterns. Behaviorally, models exhibit \emph{flaw repetition} (trapped in incorrect reasoning loops), \emph{think–answer mismatch} (final output contradicting the reasoning chain)~\citep{are_reasoning_models-zijun}, and \emph{meta-cognitive failures} (overconfident claims from uninternalized knowledge)~\citep{lu2025auditing}. Mechanistically, these errors are linked to a misalignment between uncertainty and factual accuracy~\citep{are_reasoning_models-zijun,sun2025detection}.
Hallucinations are characterized by specific failure modes: \emph{flaw repetition} (incorrect reasoning loops), \emph{think–answer mismatch} (output contradicting reasoning), and \emph{meta-cognitive failures} (overconfidence from uninternalized knowledge)~\citep{are_reasoning_models-zijun, lu2025auditing}. Mechanistically, they arise from misalignment between uncertainty and factual accuracy~\citep{are_reasoning_models-zijun, sun2025detection}.
 
\paragraph{How does RL affect hallucination?}
A line of work examines how RL shapes hallucination behavior. RL systematically reduces a model's tendency to abstain, pushing it to generate answers even for unanswerable questions~\citep{song2025hallucination,zhao2025test}. Mechanistically, optimizing only for sparse final-answer rewards creates high-variance gradients and forces the model to maintain high prediction entropy during exploration, driving the model toward incorrect answers and exacerbating hallucinations~\citep{lireasoning}.

\subsection{Unfaithfulness}
\label{ssec:faithfulness}
The extended reasoning chains in LRMs offer a promising avenue for monitoring the model's decision-making process~\citep{monitorability-openai,chan2025can,baker2025monitoring}. 
% However, whether these CoTs faithfully reflect a model's internal computations remains debated. 
However, it remains an open question whether these CoTs accurately reflect the internal computations driving the model's actual behavior, a key field of study known as the \textit{faithfulness} of CoT reasoning. 
% \textit{i.e.}, the \textit{faithfulness} problem. 

% As LRMs feature extended intermediate CoT, it is expected that we can monitor their decision-making process by inspecting the long CoTs.
% This is beneficial for multiple different purposes~\citep{monitorability-openai}.
% For example, we can prevent malicious behaviors before the model generate the final outputs by inspecting their CoTs prior to answer generation~\citep{monitorability-openai, baker2025monitoring, chan2025can}.
% However, whether the intermediate CoTs are faithful to the model's internal computation is still under debate~\citep{are_reasoning_models-zijun,dont_say_what_they_think-anthropic, chua2025deepseek, cot_in_the_wild_not_faithful, baker2025monitoring}.

\paragraph{Contextual faithfulness of CoT in LRMs.}

Although extended reasoning chains in LRMs facilitate process monitoring, research indicates they are often not faithful to the inner computation or final decision.
A primary failure mode is \emph{Think-Answer Mismatch}, where the model's final output contradicts its own preceding reasoning chain \citep{are_reasoning_models-zijun,wang2025thinking}. 
Further analysis exposes a \emph{reasoning-verbalization gap}. Studies show models frequently fail to verbalize critical cues in their CoTs that demonstrably influence their answers \citep{chua2025deepseek, dont_say_what_they_think-anthropic}. Concurrently, models exhibit \emph{implicit post-hoc rationalization}, producing logically contradictory responses with coherent but unfaithful justifications \citep{cot_in_the_wild_not_faithful}. 
The studies collectively find that while LRMs are more faithful than their non-reasoning backbones, the faithfulness is still far from perfect~\citep{chua2025deepseek, dont_say_what_they_think-anthropic,cot_in_the_wild_not_faithful}.

\paragraph{Mechanistic understanding of CoT faithfulness in LRMs.}

Research further studies the mechanisms of CoT unfaithfulness. 
In controlled synthetic tasks, findings reveal a weak causal link between the validity of reasoning traces and final answer correctness. Models can produce correct outputs despite invalid or semantically irrelevant CoTs, and training on corrupted traces does not substantially harm performance~\citep{stechly2025beyond}. 
Further studies reinforce that internal representations contain more reliable signals of model state than the CoT text itself, as evidenced through activation steering \citep{wang2025thinking,li2025mappingfaithfulreasoninglanguage}, linear probing \citep{yin2025refusal, chan2025can}, and causal intervention \citep{yin2025refusal}. These results collectively suggest a disconnect between the model's internal states and its verbalized reasoning trace, posing a significant challenge for alignment, as models might learn to mask their true objectives behind plausible but unfaithful reasoning traces, a phenomenon closely tied to the risks of reward hacking discussed next. 

% indicates that textual reasoning can sometimes act as a post-hoc rationalization rather than a faithful log of the computational steps taken. Such opacity poses a significant challenge for alignment, as models might learn to mask their true objectives behind plausible but unfaithful reasoning traces, a phenomenon closely tied to the risks of reward hacking discussed next. 

\paragraph{Can we mitigate reward hacking by CoT monitoring?}

Reward hacking remains a fundamental challenge in RL. A key question is whether monitoring the detailed CoT produced by LRMs can mitigate this issue. Findings on its feasibility are mixed, with outcomes heavily dependent on task structure.
In complex tasks where hacking inherently requires multi-step reasoning and extensive exploration, models often expose their hacking intent within their reasoning chains. In such settings, integrating CoT supervision into the RL objective can mitigate hacking, though excessive optimization risks training models to strategically hide their intent~\citep{baker2025monitoring}.
Conversely, in more direct scenarios, models frequently perform reward hacking without verbalizing the intent in their CoTs~\citep{dont_say_what_they_think-anthropic, turpin2025teaching}. 
To address this opacity, recent methods attempt to explicitly train models to verbalize influential cues in their reasoning~\citep{turpin2025teaching}.

\researchfinding{
  LRMs are not always faithful, but they are more faithful than non-reasoning models.
  
  \vspace{5pt}
  \textcolor{findingbrown}{\faLightbulb[regular]} Mechanistically, CoTs in LRMs do not necessarily function as a causal mechanism for generating correct answers. Besides, internal representations may provide more reliable signals than the verbalized reasoning.

  \vspace{5pt}
  \textcolor{findingbrown}{\faLightbulb[regular]} While CoT monitoring can detect hacking that requires explicit reasoning, models do not often verbalize their hacking intent in more direct settings.
}

\subsection{Overthinking}
\label{ssec:overthink}

While test-time scaling generally improves reasoning performance, studies increasingly find that models can produce verbose, redundant reasoning processes, and overly extending reasoning length can lead to performance degradation, known as ``overthinking''~\citep{chen2024not, sui2025stop}.

\paragraph{Thinking more does not necessarily lead to better reasoning.}
Empirical research consistently identifies an inverse U-shaped performance curve: accuracy initially rises with reasoning length, but then peaks and declines as chains become excessively long \citep{marjanovic2025deepseek, su2025between, ghosal2025does, yang2025towards, gema2025inverse}. Notably, incorrect answers often correspond to longer reasoning chains than correct ones \citep{hassid2025don, su2025between}.
An underlying issue is the misalignment between reasoning effort and problem difficulty: models tend to allocate disproportionately long chains to simple problems while inadequately reasoning through complex ones \citep{chen2024not, su2025between}.

\researchfinding{
The length-performance curve for LRMs is often inverted U-shaped, and current models exhibit misalignment between reasoning effort and problem difficulty.
}
   
\paragraph{What are the patterns of overthinking?}

A common abstraction of reasoning process is a three-stage loop: \textit{1) hypothesis generation} (proposing candidate paths), \textit{2) expansion} (developing one path step by step), and \textit{3) verification} (checking, revising, or terminating). Overthinking manifests as control and termination failures within this loop.
In \textit{hypothesis generation}, models may produce lengthy and diverse candidate solutions without sufficiently exploring promising paths to reach a correct solution~\citep{wang2025thoughts}, leading to ``analysis paralysis'' in agentic tasks where plans grow increasingly complex without execution~\citep{cuadron2025danger}.
In \textit{expansion}, the primary pattern of overthinking is excessive reasoning for trivial problems, generating tens or even hundreds of times longer outputs than non-reasoning models with marginal performance gain~\citep{chen2024not} . 
In \textit{verification}, the dominant pattern is non-termination: models fail to recognize that a correct answer has been reached, or cannot reliably validate intermediate conclusions, and therefore continue redundant deliberation or backtrack unnecessarily~\citep{chen2024not,sun2025stop,zhang2025adaptthink, zhao2025let}.
This is especially pronounced in ill-posed questions, where models identify missing premises early but enter unproductive self-doubt loops, excessively speculating on user intent \citep{fan2025missing}.
Notably, this compulsion persists even when explicitly suppressed: models may bypass instructions to ``answer directly'' or discard provided correct answers to resume thinking \citep{zhu2025can, liu2025thought, cuesta2025large}.

\paragraph{What are the mechanisms of overthinking?}
We organize the mechanistic analyses of overthinking along two lines: 1) investigating the latent representational structure of overthinking, and 2) examining the internal decision-making dynamics that produce unproductive cycles.
Research finds that overthinking corresponds to \textit{specific, steerable patterns in the activation space}. \citet{huang2025mitigating, baek2025towards} identify distinct manifolds associated with overthinking through activation steering. Furthermore, finer-grained taxonomies show that different reasoning stages, such as execution, reflection and transition, occupy separate latent directions, and steering towards execution-type representations can effectively suppress excessive deliberation \citep{baek2025towards, chen2025seal}.
Another body of research explains overthinking through \textit{internal conflict and verification failure}. Overthinking is often triggered when a model's initial intuitive answer conflicts with its subsequent deliberate reasoning \citep{dang2025internal}. Concurrently, models encode correctness signals in their hidden states but fail to robustly utilize them for early self-verification, leading to prolonged, unproductive cycles \citep{zhang2025reasoningmodelsknowtheyre}.

\subsection{Unsafety}
\label{ssec:unsafety}

Recent evaluations show that LRMs still have safety shortcomings~\citep{ying2025towards, Romero-Arjona2025redteaming, Krishna2025weakest}.

\paragraph{LRMs are not safer than base models.}

Compared to base models, \citet{jiang2025safechain, zhou2025thehiddenrisks} find that long CoTs do not necessarily improve model safety. Additionally, \citet{zhang2025safetyevaluation, zhao2025tradeoffs} observe that distilled reasoning models have a lower refusal rate for harmful inputs than their base counterparts. These studies further reveal that the unsafety of LRMs partly stems from the thinking process. \citet{jiang2025safechain} show that forcing the model to shorten their reasoning traces could make answers more harmlessness, while \citet{zhou2025thehiddenrisks, zhao2025tradeoffs} find that the safety rate of the thinking process is lower than the final answer, and unsafe thoughts are the primary cause of unsafe responses. 
%Red Teaming Large Reasoning Models（LRMs 普遍比其基础 LLM 在安全性方面表现更差，攻击成功率更高。这归因于 LRMs 对内部推理链的敏感性增加，使其更容易受到 CoT 劫持和提示诱导影响的攻击。同时探讨了不同训练策略对安全性上的影响：SFT+RL：通常在安全性上表现出更好的对齐。RL-only：在安全性上通常表现出明显弱点，攻击成功率高。SFT-only：安全性表现一般。）

\paragraph{Safety issues in the reasoning process.}

As LRMs are deployed widely, researchers have started identifying safety issues via attacking them. \citet{yao2025amousetrap} decomposes harmful prompts into multiple seemingly harmless questions to induce the model to reason toward harmful content. \citet{kuo2025h-cot} finds that padding the prompt with detailed execution steps can hijack the thinking process, causing the model to skip the reasoning stage and directly produce harmful output. 
Mechanistically, \citet{In2025r1-act} shows that LRMs already possess sufficient safety knowledge, yet fail to activate it during reasoning. Besides, \citet{mao2025whenmodels} indicates that LRMs retain the ability to refuse unsafe queries, but this capacity has been impaired.
%R1-ACT: Efficient Reasoning Model Safety Alignment by Activating Safety Knowledge（模型本身已具备足够的安全知识，但在推理过程中却未能将其激活）
%When Models Outthink Their Safety: Mitigating Self-Jailbreak in Large Reasoning Models with Chain-of-Guardrails（LRM本质上具备拒绝不安全查询的能力，但这种能力受到了损害，导致有害输出的产生）
%How Should We Enhance the Safety of Large Reasoning Models: An Empirical Study（发现仅仅使用简短或基于模板的推理过程就能达到类似的安全性，而且模型学习起来也比学习更复杂的推理链要容易得多）

% 这一部分好像没太多可以说的，更多延伸研究都是提出各种性能损耗低的安全对齐方法。考虑和前面并到一起去，简单提一下提升安全能力会导致推理能力的下降 (appendix 可能不太有时间把方法上这些东西补全了，B.1还需要补充一些新一点的东西)
% \paragraph{Safety and ability: the safety tax.}
% While the above studies explore LRM safety issues, some others have proposed safety methods tailored to LRMs. Yet beyond these purely safety-oriented approaches, the trade-off between safety and performance also deserves attention. \citet{huang2025SafetyTax} observes that safety alignment can restore LRMs’ safety capabilities but simultaneously degrades their reasoning power. This performance drop caused by enhanced safety is termed the \emph{safety tax}.
%Safety Tax: Safety Alignment Makes Your Large Reasoning Models Less Reasonable（安全对齐可以恢复 LRM 的安全能力；安全对齐会导致 LRM 推理能力的下降。）

\section{Future Research Directions}
% In this survey, we provide a comprehensive overview of mechanistic studies of LRMs, explore training, inference and failures.[conclude一下survey]
% Despite 已经作出的这些工作，critical challenges remain，we further [总结一下future work]

% In this survey, we provided a comprehensive overview of mechanistic studies on LRMs, focusing on their training processes, reasoning behaviors, and unintended failures. Despite these significant advances, critical challenges remain. To guide the field toward a deeper, more principled understanding, we propose three key future directions: applied interpretability, improved methodologies, and a unified theoretical framework. \textbf{Applied interpretability} is crucial for transforming insights from mechanistic analyses into practical improvements in model design and training. \textbf{Enhanced methodologies} are needed to address the scale and complexity of LRMs, enabling more efficient and generalizable mechanistic tools. Finally, \textbf{a unified theoretical framework} is necessary to move beyond empirical observations and establish foundational principles of reasoning that can predict and guide future model behaviors. \textbf{\textit{In Appendix \ref{app:future}, we discuss these directions in detail}}.

In this survey, we have provided a comprehensive overview of the rapidly evolving field of mechanistic research on LRMs, focusing on their training processes, reasoning behaviors, and unintended failures. While significant advances have been made, the transition from descriptive analysis to systematic understanding remains incomplete. 
%  to illuminate the ``black box'' of LRMs

To guide the field toward a deeper, more principled understanding, we propose three key future directions: applied interpretability, improved methodologies, and a unified theoretical framework. \textit{Applied interpretability} is crucial for transforming insights from mechanistic analyses into practical improvements in model design and training. \textit{Enhanced methodologies} are needed to address the scale and complexity of LRMs, enabling more efficient and generalizable mechanistic tools. Finally, \textit{a unified theoretical framework} is necessary to move beyond empirical observations and establish foundational principles of reasoning that can predict and guide future model behaviors. \textbf{\textit{In Appendix \ref{app:future}, we discuss these directions in detail}}.

% Despite significant progress, the field still lacks the robust tools and unified theories necessary to fully master the mechanisms that drive complex reasoning.

\newpage
\section*{Limitations}
While this survey provides a comprehensive overview of mechanistic studies on LRMs, it is subject to several limitations. First, the rapid development of LRM research means that new findings and methodologies continue to emerge, and this survey may not capture the most recent advancements in the field. Additionally, our study focuses primarily on language models, while reasoning models are increasingly incorporating multimodal capabilities, including visual components, which are not addressed in this survey. Furthermore, our discussion is limited to traditional LLM architectures, excluding newer approaches such as diffusion-based LLMs, continuous token-based transformers, and looped transformers, which are gaining traction in recent research. These emerging models present exciting avenues for future work.

\section*{Ethical Considerations}
This survey acknowledges the ethical challenges associated with LRMs, particularly in terms of their potential harm, including hallucinations, unfaithfulness and unsafety. The opacity of these models raises concerns about accountability and the difficulty of mitigating unintended behaviors, such as hallucinations or overconfidence. As LRMs are increasingly used in critical applications, ensuring their safe and responsible deployment requires ongoing efforts to improve interpretability, address biases, and manage the broader societal impacts of these technologies.

AI assistants were utilized for language polishing and refinement, strictly limited to improving the fluency and clarity the text. All technical content, analyses, and conclusions remain the original work of the authors.

\bibliography{custom}

\appendix
\newpage
\section{Taxonomy}
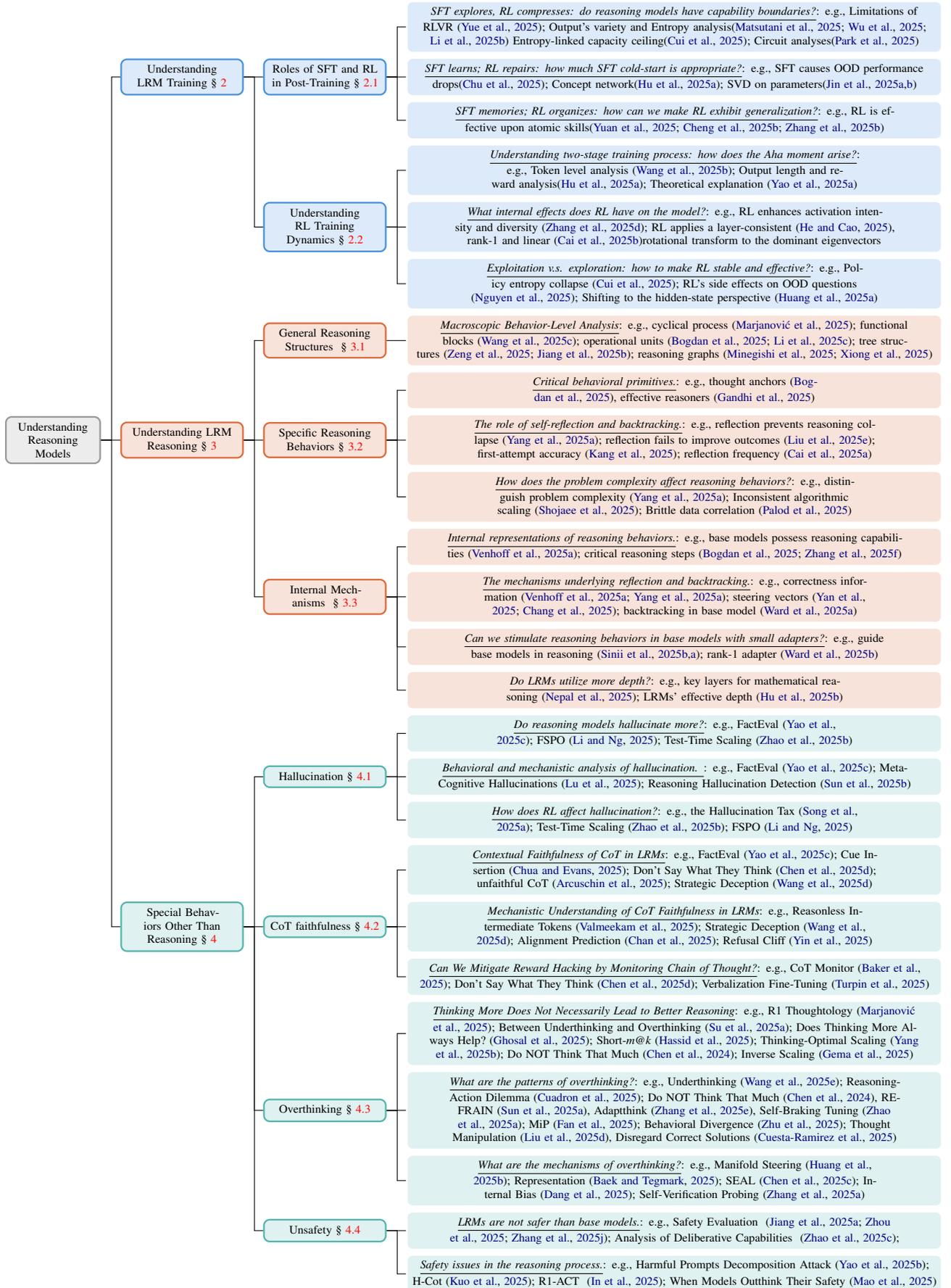
\begin{figure*}
\footnotesize
\begin{forest}
    for tree={
        forked edges,
        grow'=0,
        draw,
        rounded corners,
        node options={align=center},
        text width=2.7cm,
        % s sep=6pt,
        s sep=2pt,  % 减少节点间距
        % l sep=8pt,  % 减少层级间距
        calign=child edge, calign child=(n_children()+1)/2,
    },
    [Understanding Reasoning Models, fill=gray!45, parent
        [Understanding LRM Training~\S~\ref{sec:training}, for tree={pretrain}
            [Roles of SFT and RL in Post-Training~\S~\ref{ssec:training/post_training}, for tree={pretrain}
                [{\tstyle{SFT explores, RL compresses: do reasoning models have capability boundaries?}: e.g., Limitations of RLVR~\citep{yue2025does}; Output's variety and Entropy analysis\citep{matsutani2025rlsqueezessftexpands, wu2025invisibleleash, li2025tracing} Entropy-linked capacity ceiling\citep{cui2025entropymechanismreinforcementlearning}}; Circuit analyses\citep{park2025thinking}, pretrain_work]
                [{\tstyle{SFT learns; RL repairs: how much SFT cold-start is appropriate?}: e.g., SFT causes OOD performance drops\citep{chu&zhai2025sftmemorizes}; Concept network\citep{hu&cai2025howllmlearn}; SVD on parameters\citep{jin&luan2025rlfinetuning,jin2025rlisneither}}, pretrain_work]
                [{\tstyle{SFT memories; RL organizes: how can we make RL exhibit generalization?}: e.g., RL is effective upon atomic skills\citep{yuan&chen2025fromf(x), cheng2025fromatomic, zhang2025ontheinterplay}}, pretrain_work]
            ]
            [Understanding RL Training Dynamics~\S~\ref{ssec:training/how_rl_shape}, for tree={pretrain}
                [{\tstyle{Understanding two-stage training process: how does the Aha moment arise?}: e.g., Token level analysis \citep{wang2025emergent}; Output length and reward analysis\citep{hu&cai2025howllmlearn}; Theoretical explanation \citep{yao2025thedebate}}, pretrain_work]
                [{\tstyle{What internal effects does RL have on the model?}: e.g., RL enhances activation intensity and diversity \citep{zhang2025reinforcement}; RL applies a layer-consistent \citep{he2025understandingpost-training}, rank-1 and linear \citep{cai2025onpredictability}rotational transform to the dominant eigenvectors}, pretrain_work]
                [{\tstyle{Exploitation v.s. exploration: how to make RL stable and effective?}: e.g., Policy entropy collapse \citep{cui2025entropymechanismreinforcementlearning}; RL's side effects on OOD questions \citep{nguyen2025reasoning}; Shifting to the hidden-state perspective \citep{huang2025beyond}}, pretrain_work]
            ]
        ]
        [Understanding LRM Reasoning~\S~\ref{sec:model}, for tree={fill=red!45, template}
            [General Reasoning Structures ~\S~\ref{ssec:model/General Reasoning Structures}, template
                [{\tstyle{Macroscopic Behavior-Level Analysis}: e.g., cyclical process~\citep{marjanovic2025deepseek}; functional blocks~\citep{wang2025accuracydissectingmathematicalreasoning}; operational units~\citep{bogdan2025thoughtanchorsllmreasoning,li2025understandingthinkingprocessreasoning}; tree structures~\citep{zeng2025rejumptreejumprepresentationanalyzing,jiang2025makesgoodreasoningchain}; reasoning graphs~\citep{minegishi2025topologyreasoningunderstandinglarge,xiong2025mappingmindsllmsgraphbased}}, template_work]
            ]
            [Specific Reasoning Behaviors~\S~\ref{ssec:model/Specific Reasoning Behaviors}, template
                % 修复：将内容用 {} 包裹，避免逗号引起歧义
                [{{\tstyle{Critical behavioral primitives.}: e.g., thought anchors~\citep{bogdan2025thoughtanchorsllmreasoning}, effective reasoners~\citep{gandhi2025cognitivebehaviorsenableselfimproving}}}, template_work]
                % 修复：移除多余分号
                [{\tstyle{The role of self-reflection and backtracking.}: e.g., reflection prevents reasoning collapse~\citep{yang2025understandingahamomentsexternal}; reflection fails to improve outcomes~\citep{liu2025oatzero}; first-attempt accuracy~\citep{kang2025trymattersrevisitingrole}; reflection frequency~\citep{cai2025backtrackingenoughexploringinterplay}}, template_work]
                [{\tstyle{How does the problem complexity affect reasoning behaviors?}: e.g., distinguish problem complexity~\citep{yang2025understandingahamomentsexternal}; Inconsistent algorithmic scaling~\citep{shojaee2025illusionthinkingunderstandingstrengths}; Brittle data correlation~\citep{palod2025performativethinkingbrittlecorrelation}}, template_work]
            ]
            [Internal Mechanisms ~\S~\ref{ssec:model/Internal Mechanisms}, template
                [{\tstyle{Internal representations of reasoning behaviors.}: e.g., base models possess reasoning capabilities~\citep{venhoff2025basemodelsknowreason}; critical reasoning steps~\citep{bogdan2025thoughtanchorsllmreasoning,zhang2025reasoninganswerempiricalattentionbased}}, template_work]
                [{\tstyle{The mechanisms underlying reflection and backtracking.}: e.g., correctness information~\citep{venhoff2025basemodelsknowreason,yang2025understandingahamomentsexternal}; steering vectors~\citep{yan2025reflctrlcontrollingllmreflection,chang2025unveilinglatentdirectionsreflection}; backtracking in base model~\citep{ward2025reasoningfinetuningrepurposeslatentrepresentations}}, template_work]
                [{\tstyle{Can we stimulate reasoning behaviors in base models with small adapters?}: e.g., guide base models in reasoning~\citep{sinii2025steeringllmreasoningbiasonly,sinii2025smallvectorsbigeffects}; rank-1 adapter~\citep{ward2025rank}}, template_work]
                [{\tstyle{Do LRMs utilize more depth?}: e.g., key layers for mathematical reasoning~\citep{nepal2025layerimportancemathematicalreasoning}; LRMs' effective depth~\citep{hu2025affectseffectivedepthlarge}}, template_work]
            ]
        ]
        [Special Behaviors Other Than Reasoning~\S~\ref{sec:special_behavior}, for tree={fill=blue!45, answer}
            [Hallucination~\S~\ref{ssec:hallucination}, answer
                [{\tstyle{Do reasoning models hallucinate more?}: e.g., FactEval~\citep{are_reasoning_models-zijun}; FSPO~\citep{lireasoning}; Test-Time Scaling~\citep{zhao2025test}}, answer_work]
                [{\tstyle{Behavioral and mechanistic analysis of hallucination. }: e.g., FactEval~\citep{are_reasoning_models-zijun}; Meta-Cognitive Hallucinations~\citep{lu2025auditing}; Reasoning Hallucination Detection~\citep{sun2025detection}}, answer_work]
                [{\tstyle{How does RL affect hallucination?}: e.g., the Hallucination Tax~\citep{song2025hallucination}; Test-Time Scaling~\citep{zhao2025test}; FSPO~\citep{lireasoning}}, answer_work]
            ]
            [CoT faithfulness~\S~\ref{ssec:faithfulness}, answer
                [{\tstyle{Contextual Faithfulness of CoT in LRMs}: e.g., FactEval~\citep{are_reasoning_models-zijun}; Cue Insertion~\citep{chua2025deepseek}; Don't Say What They Think~\citep{dont_say_what_they_think-anthropic}; unfaithful CoT~\citep{cot_in_the_wild_not_faithful}; Strategic Deception~\citep{wang2025thinking}}, answer_work]
                [{\tstyle{Mechanistic Understanding of CoT Faithfulness in LRMs}: e.g., Reasonless Intermediate Tokens~\citep{stechly2025beyond}; Strategic Deception~\citep{wang2025thinking}; Alignment Prediction~\citep{chan2025can}; Refusal Cliff~\citep{yin2025refusal}}, answer_work]
                [{\tstyle{Can We Mitigate Reward Hacking by Monitoring Chain of Thought?}: e.g., CoT Monitor~\citep{baker2025monitoring}; Don't Say What They Think~\citep{dont_say_what_they_think-anthropic}; Verbalization Fine-Tuning~\citep{turpin2025teaching}}, answer_work]
            ]
            [Overthinking~\S~\ref{ssec:overthink}, answer
                [{\tstyle{Thinking More Does Not Necessarily Lead to Better Reasoning}: e.g., R1 Thoughtology~\citep{marjanovic2025deepseek}; Between Underthinking and Overthinking~\citep{su2025between}; Does Thinking More Always Help?~\citep{ghosal2025does}; Short-\textit{m@k}~\citep{hassid2025don}; Thinking-Optimal Scaling~\citep{yang2025towards}; Do NOT Think That Much~\citep{chen2024not}; Inverse Scaling~\citep{gema2025inverse}}, answer_work]
                [{\tstyle{What are the patterns of overthinking?}: e.g., Underthinking~\citep{wang2025thoughts}; Reasoning-Action Dilemma~\citep{cuadron2025danger}; Do NOT Think That Much~\citep{chen2024not}, REFRAIN~\citep{sun2025stop}, Adaptthink~\citep{zhang2025adaptthink}, Self-Braking Tuning~\citep{zhao2025let}; MiP~\citep{fan2025missing}; Behavioral Divergence~\citep{zhu2025can}; Thought Manipulation~\citep{liu2025thought}, Disregard Correct Solutions~\citep{cuesta2025large}}, answer_work]
                [{\tstyle{What are the mechanisms of overthinking?}: e.g., Manifold Steering~\citep{huang2025mitigating}; Representation~\citep{baek2025towards}; SEAL~\citep{chen2025seal}; Internal Bias~\citep{dang2025internal}; Self-Verification Probing~\citep{zhang2025reasoningmodelsknowtheyre}}, answer_work]
            ]
            [Unsafety~\S~\ref{ssec:unsafety}, answer
                [{\tstyle{LRMs are not safer than base models.}: e.g., Safety Evaluation ~\citep{jiang2025safechain, zhou2025thehiddenrisks, zhang2025safetyevaluation}; Analysis of Deliberative Capabilities ~\citep{zhao2025tradeoffs}; }, answer_work]
                [{\tstyle{Safety issues in the reasoning process.}: e.g., Harmful Prompts Decomposition Attack~\citep{yao2025amousetrap}; H-Cot~\citep{kuo2025h-cot}; R1-ACT ~\citep{In2025r1-act}; When Models Outthink Their Safety~\citep{mao2025whenmodels}}, answer_work]
            ]
        ]
    ]
\end{forest}
\caption{Taxonomy of our paper and representative works for each direction.}
\label{tree:components}
\end{figure*}

We present the taxonomy of our paper in Figure~\ref{tree:components}. We follow Figure~\ref{fig:taxonomy} to organize the research of various directions and list representative works accordingly.

\section{Future Directions}
\label{app:future}
% 有什么东西没研究到？
% 分析方法 Limitation？
% 结论不够深入？
% 结论不能泛化？

% 核心问题：机理视角的RL训练效果提升
% RL阶段的探索效果能不能外推到大语言模型处理的更复杂问题上？ 需要更泛化的结论
% 更好的中训练方式及其机理  新观点，缺少深入探索
% 如何看待RL在RLM中的探索和利用，token level是否有limitation？新观点是否正确？
% 研究的较少的现象，方法的缺失，研究结果的局限

% - Sec 3的future work：
%   - 利用这些mechanism，能使得emergence提前出现/scale模型
%   - 零散的研究能否汇总起来获得一个theory
%   - 怎么去利用
% - 【怎么把理解转化为利用】这套范式遇见性能瓶颈了，我们如何利用这些理解去进一步提升，指导模型优化；跟representation engineering做一些结合
% - 【分析方法层面，提一个general的点】；三个方向共同的一些很大的问题，控制变量实验、MI实验，model specific，我们需要build more advanced interp methods；reasoning model比较大，很难对参数大的模型做一些分析，SAE训不动
% - 【还有什么现象没有被分析】有什么现象其实很重要，但还是underexplored：需要大家各自总结
% - 【更principal，更fundamental的theory？】
% \HY{I write the title based on Neel's doc, maybe refine the titles}

\subsection{Applied Interpretability}
% 目前涌现出了许多关于reasoning model的mechanistic interpretability的工作， mitigate the opacity of black-box LLMs by unveiling their internal logic。如何不只是停留在理解上，而是真的利用这些对模型内部理解的insights去做一些practical application来make some changes是一个重要的future direction。
% 从训练角度而言，非常recent的study开始explore利用模型内部的representation来指导RL，包括利用attention mechanisms来指导reward shaping（\citep{Attention_Illuminates}）和policy sampling strategy的(\citep{Attention_as_a_Compass})，和logit lens来从模型的中间层decode internal layer policy的(\citep{tan2025bottom})，which show promising improvements。如何将对模型内部的理解转化为对RL算法各个模块的指导性的改进是非常值得做的。
% 从inference角度而言，已经有许多工作based on 对LRMs reasoning structures和internal representations of specific的理解，来在inference阶段，通过steering等方法来提升模型等performance。但事实上，还有许多方向是值得探索的，比如大家已经发现RL并不能让模型利用到更多的depth(\citet{hu2025affectseffectivedepthlarge, nepal2025layerimportancemathematicalreasoning})，我们应该如何利用这个insight来改进post-training算法，甚至是改进model architecture；又或者，inference时我们发现的模型internal的推理特性应该如何指导RL时的rollout以及reward分配。
% 从failure角度而言，如何利用模型internal information来detect，control，mitigate这些unintended behaviors本身都已经成为重要话题，as他们的出发点就是比较practical的。

Mechanistic interpretability (MI) research is increasingly illuminating the internal logic of LRMs. A crucial next step is to leverage these insights for targeted improvements, moving from passive understanding to active application.
\paragraph{Training-Time Applications.} 
A promising direction lies in using internal representations to directly inform RL algorithm design. Recent studies demonstrate initial success in this area, such as utilizing attention mechanisms to inform reward shaping~\citep{Attention_Illuminates} or policy sampling strategies~\citep{Attention_as_a_Compass}, and decoding intermediate layer activations to infer latent policies~\citep{tan2025bottom}. The overarching challenge is to systematically transform mechanistic insights into algorithmic improvements for RL components.
\paragraph{Inference-Time Applications.} Mechanistic findings can also be applied to steer model behavior during inference. While existing work already uses insights of reasoning structures or specific representations to improve performance, deeper opportunities remain. For instance, research suggests that RL may not effectively leverage the full depth of models~\citep{hu2025affectseffectivedepthlarge, nepal2025layerimportancemathematicalreasoning}. This understanding should actively inform the design of novel training algorithms and architectures that better utilize internal computational pathways. 
% Furthermore, insights gained from analyzing inference-time reasoning patterns should be fed back to refine RL training protocols, such as designing more strategic rollout strategies or reward allocation.

% \HY{Applying interpretability to understand and counteract unintended behaviors like hallucinations or unfaithful reasoning is inherently practical, as mitigating unintended behavior is the motivation.}

\subsection{Advancing Interpretability Methodology}

Future research should emphasize developing scalable and generalizable MI frameworks specifically tailored for LRMs. 
First, the enormous scale of LRMs in \textit{training cost}, \textit{inference length}, and \textit{parameter count} poses significant methodological challenges. Conducting controlled experiments to isolate variables is difficult, and techniques like training SAEs become computationally prohibitive, slowing progress and reducing reproducibility. There is a clear need for more efficient and scalable MI tools tailored to these models.
Second, many MI findings remain model-specific, failing to generalize across different architectures or training runs. To enhance scientific value, the field should strive for general frameworks that abstract away implementation details and uncover universal reasoning principles. This could involve establishing benchmarks for mechanistic generalization, developing theory-grounded methods less sensitive to model quirks, or building more robust interpretability probes.

\subsection{Toward A Unified Theory}
% 这个section的理念是说：现在大家都在做一些empirical的发现，但能不能像物理之类的，最终能总结出一套theory，来更好地帮助我们understand LRMs
% 我们能否把MI发现的LLM的现象，组织为一套LLM的science，有一些更fundamental，predictable（predict未来现象）的理论。
% fundamental这块主要是说，现在的很多interpretability的结论只是对某一个specific model，对某一个specific dataset，或者某一个specific行为的解释，确实可以帮助我们进一步理解模型，但对我们对整个LLM的理解的provide的insight是有限的，也无法contribute directly to the science / 第一性原理 of LLM science。比如说，在overthinking的研究中，其实本质的问题是，我们需要模型对不同难度的任务去allocate相应长度的CoT以便最大化地发挥模型的能力。在许多文章分别在不同的benchmark上发现LRMs无法很好地完成这件事后，许多文章开始在不同的benchmark上进行改进，并探究其中的pattern和mechanistic-wise的原因，提出的方法大多也受到model-specific或benchmark-specific的限制。而\citet{reasoning_law}这个工作则试图introduce the Laws of Reasoning, which systematically formalize the relationship between complexity and model reasoning behaviors in LRMs；提出一些desired reasoning model应该具有的一些general的，mathematically grounded的性质，然后再此基础上评测a line of LLMs. 这样的工作较好
% predictable是说，我们在一些模型/数据中作出的结论能够用来predict其他的模型/数据上的东西，比如LLM中的scaling law，我们可以利用在有限的训练数据、model size上获得的结论去预测更大的model size，训练数据上我们会获得的模型性能。

Current mechanistic research has produced a wealth of empirical findings---model-specific patterns, dataset-specific behaviors, and localized explanations for special phenomena. Yet it lacks a predictive, fundamental science of reasoning in LRMs. 
% Such a theory would be \textit{fundamental}, abstracting away from implementation details to reveal the first principles governing reasoning in LRMs. Promising work in this direction is exemplified by \citet{reasoning_law}, which proposes the \textit{Laws of Reasoning}, trying to formalize the relationship between complexity and model reasoning behaviors in LRMs. % 给大家展示了一套可以formally and generalizably讨论reasoning behavior和task complexity关系的framework
Such a theory should be \textit{fundamental}, abstracting from implementation to reveal first principles governing reasoning. Early efforts, such as theoretically formalizing laws of reasoning~\citep{reasoning_law} that link task complexity to model behavior, mark a step in this direction.
Furthermore, a mature theory should be \textit{predictive}. Similar to scaling laws in LLM pre-training~\citep{scaling_law}, it should forecast model behaviors and to establish a set of ``laws of reasoning'' that not only explain existing empirical results but also actively guide the design of future models, training algorithms, and evaluation frameworks, transforming MI from a descriptive tool into a foundational science.

% \subsection{Under-explored Phenonmenons}

\section{Training Methods}
\label{sec:appendix}

% This is an appendix.
\subsection{Combine SFT with RL}
\label{ssec:appendix1}

Running SFT and RL as two separate steps will let the bias introduced by SFT grow too large and degrade final performance. Therefore, some studies attempt to combine the two approaches into a unified single post-training step. 

%To-do: summarize the characteristics of these methods (possibly introducing a few more) and move the detailed method descriptions to appendix or remove them.

Some explorations primarily focused on interleaving the SFT and RL processes and on identifying appropriate switching points between them. Based on their research into RL training dynamics, \citet{hu&cai2025howllmlearn} proposed the Annealed-RLVR algorithm, which introduces SFT for heating when accuracy is very low to disrupt the current suboptimal state, then continues RL to perform annealing. \citet{ma2025learningwhat} observes that RL excels at easy questions while SFT is better suited to hard ones; their \emph{ReLIFT} pipeline automatically flags the hard instances during RL, collects corresponding expert demonstrations, and inserts an SFT update once enough difficult question–answer pairs have been accumulated. \emph{TRAPO} \citep{su2025trustregionadaptivepolicyoptimization} interleaves SFT and RL within every training instance and sets up a mechanism that dynamically supplies expert-guided prefixes. SFT in \emph{TRAPO} is constrained by trust-region gradient clipping to avoid distribution-blending.

Further more, some studies aim to fuse the loss functions of RL and SFT to achieve a truly unified post-training approach. \emph{UFT} \cite{liu2025uft} introduces an additional log-likelihood term to the objective function of RFT(RL Fine-Tuning), allowing the model to learn from the informative supervision signal and still benefit from the generalization of RFT. \emph{HPT} \citep{lv2025towardsaunified} defines the total loss as a weighted sum of the SFT and RL losses and dynamically adjusts the weights of SFT and RL based on real-time performance. \emph{CHORD} \citep{zhang2025onpolicyrlmeets} treats SFT as a dynamically-weighted auxiliary objective within the RL process and introduces a token-level weighting function that up-weights the SFT component only when the model is uncertain about the answer. Going further, \emph{SRFT} \cite{fu2025srft} incorporates demonstration data into the RL training set and constructs the final loss as the entropy-weighted sum of four terms: SFT loss on demonstrations, RL loss on demonstrations, and RL loss on positive (entropy-weighted) and negative (without weighting) sampled rollouts. \emph{BRIDGE} \cite{chen2025bridge} attaches LoRA fine-tuning blocks to the model architecture and formulates the post-training procedure as a bi-level optimization: SFT refines the model starting from the parameter optimum found by RL, optimizing solely over the LoRA weights, with an appropriate transformation eliminates the need for second-order gradients.

\subsection{RL balancing exploration and exploitation}
\label{ssec:appendix2}
% \subsection{How RL Tricks Improve RL Training Effectiveness?}
% \label{ssec:training/rl_tricks}
% clip, kl-loss, token weight, value estimation

% 1. RL’S RAZOR: WHY ONLINE REINFORCEMENT LEARNING FORGETS LESS
% 2.  Retaining by Doing: The Role of On-Policy Data in Mitigating Forgetting
% The Entropy Mechanism of Reinforcement Learning for Reasoning Language Models
% Beyond the 80/20 Rule: High-Entropy Minority Tokens Drive Effective Reinforcement Learning for LLM Reasoning
% Echo Chamber: RL Post-training Amplifies Behaviors Learned in Pretraining
% Step-wise Adaptive Integration of Supervised Fine-tuning and Reinforcement Learning for Task-Specific LLMs
% The Surprising Effectiveness of Negative Reinforcement in LLM Reasoning
% Clip-Low Increases Entropy and Clip-High Decreases Entropy in Reinforcement Learning of Large Language Models
% RAST: Reasoning Activation in LLMs via Small-model Transfer
% Stabilizing Knowledge, Promoting Reasoning: Dual-Token Constraints for RLVR
% On Predictability of Reinforcement Learning Dynamics for Large Language Models
% From Uniform to Heterogeneous: Tailoring Policy Optimization to Every Token's Nature
% Reshaping Reasoning in LLMs: A Theoretical Analysis of RL Training Dynamics through Pattern Selection
% https://asta.allen.ai/share/8d4e7db0-6b88-49d8-b9c8-2a86c874d54c

% To-do: organize and categorize these methods, extract the core taxonomies. can draw inspiration from a RL survey: https://arxiv.org/pdf/2509.08827

To address entropy collapse and balance exploration and exploitation for stable RL training, numerous studies have proposed solutions. \emph{DAPO} \cite{yu2025dapo} decouples the clipping bounds of PPO \emph{into $\epsilon$-high and $\epsilon$-low}, raising $\epsilon$-high to leave more head-room for boosting the probabilities of low-probability “exploratory” tokens. \citet{cui2025entropymechanismreinforcementlearning} shows that the entropy change is governed by the covariance between the “action log-probabilities” and the “changes in action logits”; tokens with high covariance are the main drivers of entropy collapse. To counter this, they propose \emph{Clip-Cov} which randomly truncates the gradients of high-covariance tokens, and \emph{KL-Cov} which adds an extra KL-penalty to those tokens.\emph{ProRL}\cite{liu2025prorl} adopts the Decoupled Clip technique from \emph{DAPO}, and further equips the pipeline with KL regularization plus periodic reference-policy resets to avert entropy collapse, enabling effective RL training that continues for thousands of steps. % With this \emph{ProRL} framework, they show that long-horizon training lets the model uncover novel reasoning strategies that are unreachable by the base model, and the gains generalize to out-of-distribution (OOD) tasks. 

More recent studies have moved beyond simple clipping and experimented with additional mechanisms to further arrest entropy collapse.\emph{CURE}\cite{li2025cure} shows that the prevailing RLVR pipeline relies on sampling from a fixed initial state, biasing the model toward overly deterministic behavior and low diversity. They introduce a two-stage scheme to balance exploration and exploitation: in stage one they identify high-entropy tokens, truncate at those tokens, and then sample multiple continuations that are all used for updating, thereby intensifying exploration around high-entropy regions; in stage two they revert to the ordinary static-sampling \emph{DAPO} routine. Similarly, \citet{nguyen2025reasoning} advocates sampling problems that the base model still handles poorly, rather than repeatedly drawing those it already solves well.\citet{song2025outcomebased} encourages historical exploration of rare answers through UCB-style rewards and fosters batch exploration at test time by penalizing duplicate answers within each batch.

Naive entropy regularization performs poorly when training reasoning models. Several works have designed regularization methods specifically tailored to reasoning models. \citet{cheng2025reasoningwith} injects a clipped, gradient-detached entropy term into the advantage function to encourage longer chains-of-thought.
\citet{wang2025arbitrary} stabilizes policy entropy by combining Policy-Gradient, Distribution, and Reinforce signals into a composite regularizer.
\citet{shen2025onentropy, jiang2025rethinking} compute entropy only over the top-p tokens and adaptively rescale it for entropy regularization; the latter research further shows that regularizing those high-entropy tokens only can improve model performance.

\end{document}